# Machine Learning-based Estimation of Respiratory Fluctuations in a Healthy Adult Population using BOLD fMRI and Head Motion Parameters[*]

Abdoljalil Addeh [1-3,5], Fernando Vega [1-3,5], Rebecca J Williams[6]
G. Bruce Pike,[3-5] M. Ethan MacDonald [1-3,5]

[1] Department of Biomedical Engineering, Schulich School of Engineering, University of Calgary, Canada
[2] Department of Electrical & Software Engineering, Schulich School of Engineering, University of Calgary, Canada
[3] Department of Radiology, Cumming School of Medicine, University of Calgary, Canada
[4] Department of Clinical Neurosciences, Cumming School of Medicine, University of Calgary, Canada
[5] Hotchkiss Brain Institute, Cumming School of Medicine, University of Calgary, Canada
[6] Faculty of Health, Charles Darwin University, Australia

**Synopsis:**
**Motivation**: In many fMRI studies, respiratory signals are often missing or of poor quality. Therefore, it could be highly beneficial to have a tool to extract respiratory variation (RV) waveforms directly from fMRI data without the need for peripheral recording devices.
**Goal**: Investigate the hypothesis that head motion parameters contain valuable information regarding respiratory patter, which can help machine learning algorithms estimate the RV waveform.
**Approach:** This study proposes a CNN model for reconstruction of RV waveforms using head motion parameters and BOLD signals.
**Results:** This study showed that combining head motion parameters with BOLD signals enhances RV waveform estimation.
**Impact**
It is expected that application of the proposed method will lower the cost of fMRI studies, reduce complexity, and decrease the burden on participants as they will not be required to wear a respiratory bellows.

**Introduction**

Acquisition of clean external respiratory data during functional magnetic resonance imaging (fMRI) to remove the effect of low-frequency respiratory variation is not always possible [1]. Several machine learning-based approaches have been developed recently that utilize the information contained within blood oxygen level-dependent (BOLD) signals to estimate respiratory variations (RV) waveforms [2-4]. These methods showed promising results when tested on the Human Connectome Project in Young Adults (HCP-YA) dataset but faced limitations in complete RV signal reconstruction due to edge-effects. Respiration can also influence head motion in fMRI. Recent studies have shown that respiration generates real and pseudomotion of the head at the respiratory rate (~0.3 Hz for healthy adults) [5]. Thus, there may be a potential use for head motion parameters in RV estimation using machine learning approaches. The aim of this study is to investigate the hypothesis that head motion parameters contain valuable information regarding the different respiratory events, which can help machine learning algorithms estimate the RV waveform.

**Method**

We utilized 900 resting-state fMRI scans from the HCP-YA dataset. Fig.1 demonstrates how current respiration affects subsequent BOLD signals and head motion using a gray plot of BOLD signals [6]. In addition, current breathing depth and rate are dependent upon previous breaths [7]. Fig. 2 displays the power spectra derived from motion parameters. Notable oscillations around 0.3 Hz and 0.12 Hz, especially in the phase encoding direction, are evident. These oscillations are attributed to both the physical movement of the head due to breathing and the pseudomotion artifact, which arises from changes in lung volume and subsequent shift in the $B_0$ field [5].

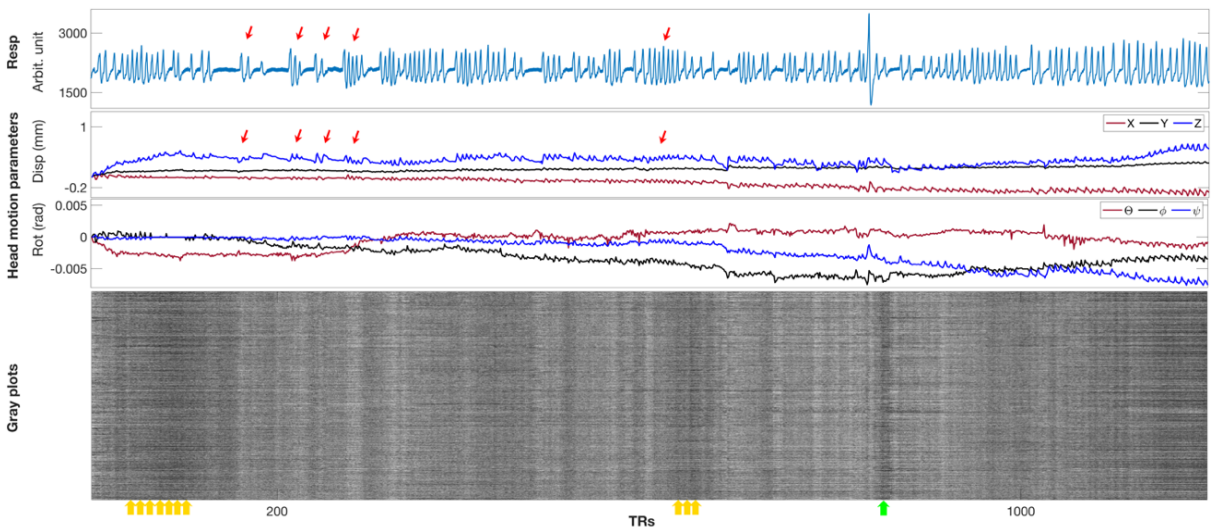

**Figure 1**. Illustration of respiratory events impact on BOLD signals and subject's head motion parameters. Vertical black bands after each single isolated deep breath (shown by a green arrow) or burst of deep breaths (shown by yellow arrows) reflect the BOLD signal decrease. During spontaneous breath-holding, the subject has a little head motion, and during deep breaths, the subject shows a high head motion (shown by red arrows).

The proposed method for reconstructing RV waveforms employs three 1D-CNNs in the temporal dimension of the BOLD time series and head motion parameters to reconstruct RVs (Fig. 3). To decrease computational complexity, the average BOLD signals from 90 functional regions of interest (ROI) [8] was used. The RV is defined as the standard deviation of the respiratory waveform within a six second sliding window [9]. The size of the moving window is selected as 65 in the proposed method. Therefore, each input had a size of [65 × 96], where 65 is the window size, and 96 is the number of regions of interest and six head motion parameters. In the proposed methods, Method 1 estimates the RV at the beginning, Method 2 at the middle, and Method 3 at the end of the scan.

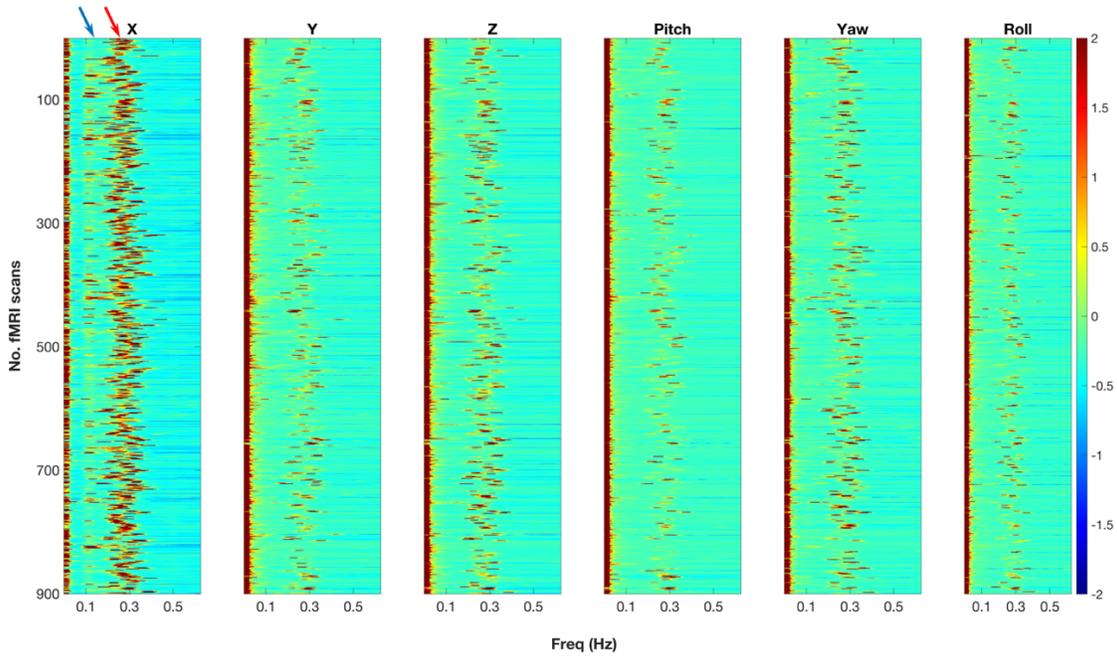

**Figure 2.** Power spectral density of motion parameters for HCP-YA dataset. In HCP-YA project, the phase encoding direction is Left→Right and Right → Left. The respiration creates head pseudomotion at frequency of ~0.3 Hz as shown by red arrow, which consistent with the normal breathing rate of this age group. Blue arrow shows the frequency band that deep breaths occur.

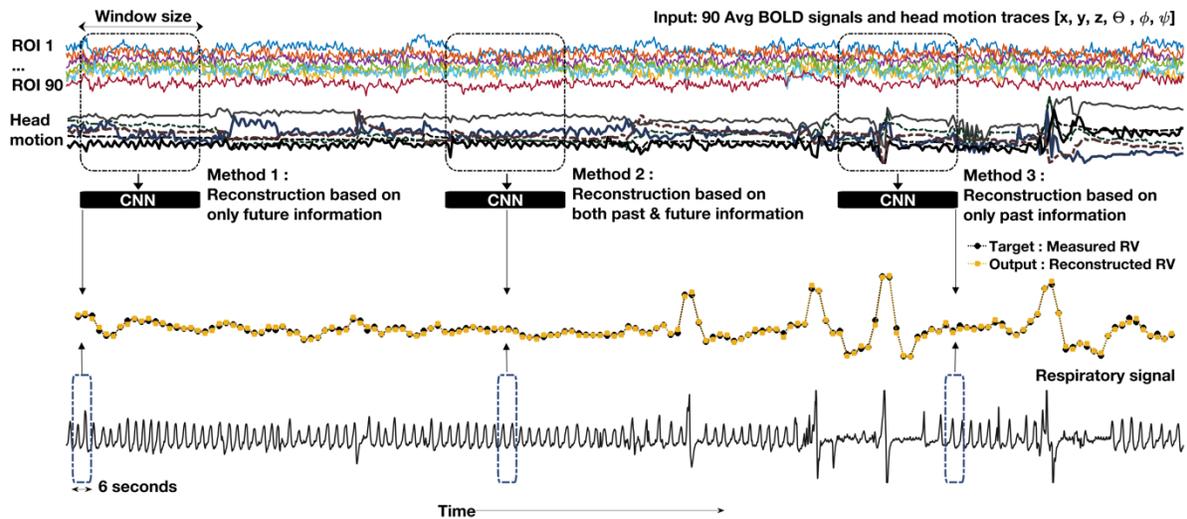

**Figure 3.** Inputs and outputs in the proposed method. The proposed method uses both BOLD signals and head motion parameters as the input of model to reconstruct RV waveforms. Method 1 uses the information in the future BOLD signals, Method 2 takes advantage of both sides, and Method 3 use only the past information. For windows size 65 and fMRI data with 1200 volumes as an example, Method 1 will estimate $[RV]_{1:32 \times 1}$, Method 2 will estimate $[RV]_{33:1168 \times 1}$, and Method 3 will estimate $[RV]_{1167:1200 \times 1}$.

**Results**

In this section, each reported value or plot represents results over the unseen test data in 10-fold cross-validation in terms of mean absolute error (MAE), mean square error (MSE), Pearson correlation, and dynamic time warping (DTW). The model's predictive accuracy was quantified using the Mean Absolute Error (MAE), which captures the average magnitude of prediction errors. Additionally, the Mean Squared Error (MSE) was utilized to evaluate the influence of larger errors, which is crucial for identifying significant respiratory events in fMRI data. The Pearson correlation coefficient was applied to measure the linear relationship between the predicted and actual respiratory variation signals, showcasing the model's ability to follow the general trend of the signal. The Dynamic Time Warping (DTW) distance metric was also used to assess the temporal alignment of the predicted time-series with the actual data, an important factor for sequential data like respiratory cycles. A lower DTW distance suggests greater precision in the model's temporal predictions. Utilizing these diverse metrics provides a thorough validation of the model's ability to accurately reconstruct respiratory variation, which is vital for dependable analysis of fMRI data.

Figure 4 illustrates the efficacy of the proposed method compared to the prior approach [4], which utilized only BOLD signals as input for the machine learning model. The trained network can reconstruct the RV timeseries with high accuracy, especially when there are big changes in RV value. We noted a similarity between the primary frequencies of the respiratory signal and the head motion parameters, as illustrated in Figure 2. This fact allows the CNN to extract critical information about the breathing rate, enhancing RV reconstruction accuracy. Figure 4, highlighted by orange arrows, demonstrates instances where the RV, reconstructed using both BOLD signals and head motion parameters, closely matches the measured RV during significant respiratory events. Conversely, when a subject maintains consistent breathing rhythm and depth, the CNN struggles to discern pertinent details for RV reconstruction. However, inputting breathing rate information via head motion parameters enhances accuracy, as depicted by the green arrows in Figure 4.

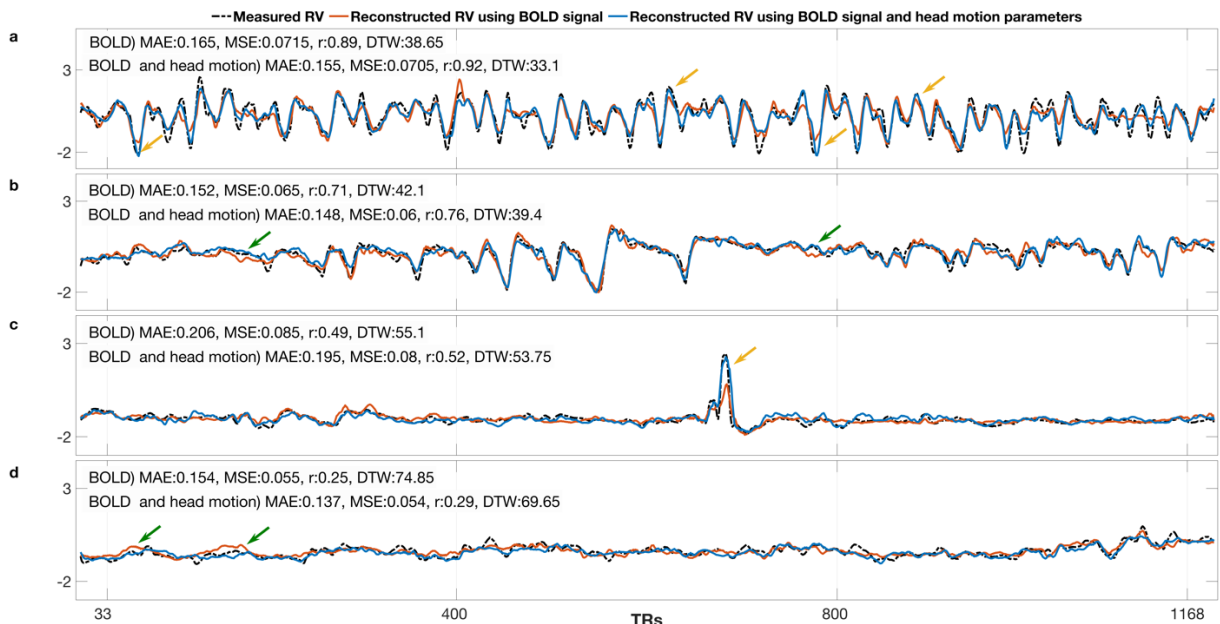

**Figure 4.** The performance of the proposed method for RV reconstruction on the test dataset. RV timepoints between TRs 33 and 1168 (shown by green arrows), are reconstructed using Method 2, initial RV timepoint (before TR 33) are reconstructed using Method 1, and last RV timepoints (after TR 1168) are reconstructed using Method 3.

Figure 5 shows the performance of the CNN in terms of MAE, MSE, r, and DTW. Across all evaluated metrics, the results highlight the benefits of integrating head motion parameters with BOLD signals in the input dataset. Statistical analysis conducted with the Friedman test indicates a statistically significant difference between the two methodologies ($p < 0.01$).

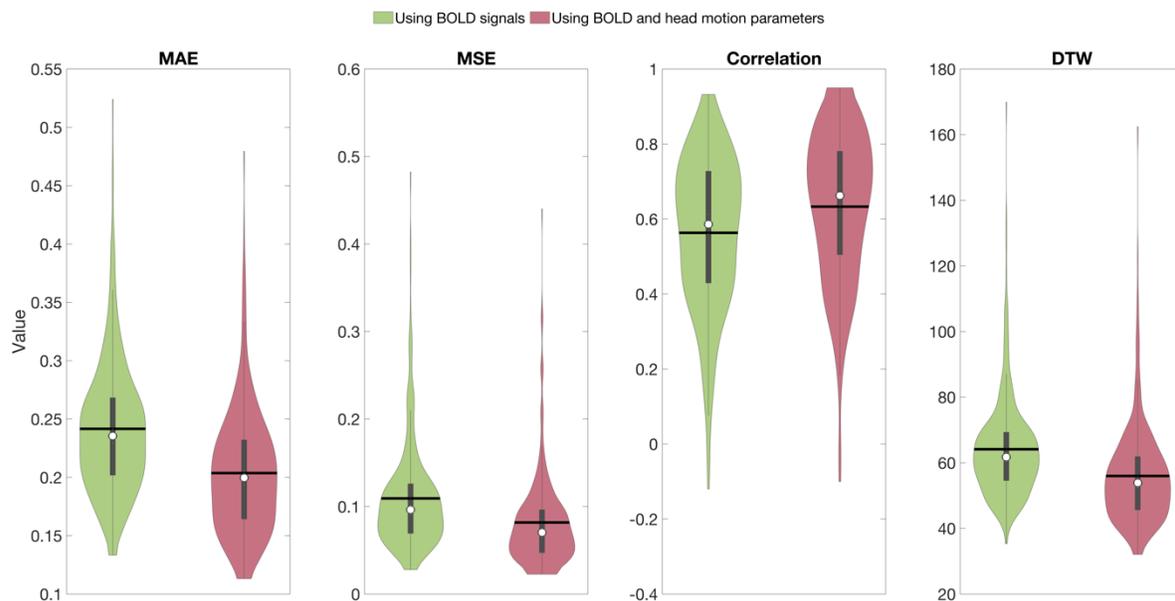

**Figure 5.** Illustration of the proposed method's performance on test samples using violin plots and studying the impact of model's input on reconstruction accuracy. A black horizontal line indicates the Mean value, while a white circle and two triangles represent the Median value and notches, respectively. Head motion parameters improved the model's performance in terms of all metrics.

## Conclusion

This study showed that combining head motion parameters with BOLD signals enhances RV waveform estimation compared to using only BOLD signals. Head motion parameters' high-frequency components indicate the primary breathing rate, while the low-frequency components detect deep breaths during fMRI. Beyond using the reconstructed RV series to correct physiological confounds from fMRI, it offers insights into subjects' breathing consistency across scans, assisting in fMRI data interpretation. Our method enriches fMRI studies without needing respiratory data, positively influencing data quality, interpretation, retention, statistical power, costs, participant burden, and adding physiological context to existing fMRI datasets.

## Acknowledgements:

The authors would like to thank the University of Calgary, in particular the Schulich School of Engineering and Departments of Biomedical Engineering and Electrical & Software Engineering; the Cumming School of Medicine and the Departments of Radiology and Clinical Neurosciences; as well as the Hotchkiss Brain Institute, Research Computing Services and the Digital Alliance of Canada for providing resources. The authors would like to thank the Human Connectome Project for making the data available. JA – is funded in part from a graduate scholarship from the Natural Sciences and Engineering Research Council Brain Create. GBP acknowledges support from the Campus Alberta Innovates Chair program, the Canadian Institutes for Health Research (FDN-143290), and the Natural Sciences and Engineering Research Council (RGPIN-03880). MEM acknowledges support from Start-up funding at UCalgary and a Natural Sciences and Engineering Research Council Discovery Grant (RGPIN-03552) and Early Career Researcher Supplement (DGECR-00124).